\documentclass[10pt,twocolumn,letterpaper]{article}

\usepackage{cvpr}              

\usepackage[dvipsnames]{xcolor}

\definecolor{cvprblue}{rgb}{0.21,0.49,0.74}
\usepackage[pagebackref,breaklinks,colorlinks,citecolor=cvprblue]{hyperref}

\def\paperID{*****} 
\def\confName{CVPR}
\def\confYear{2024}

\title{Leveraging Perceptual Scores for Dataset Pruning in Computer Vision Tasks}

\author{Raghavendra Singh\\
Ashoka University \thanks{Visiting Faculty, CS}\\
Delhi, India\\
{\tt\small raghavendra.singh@ashoka.edu.in}
}

\begin{document}
\maketitle
\begin{abstract}
In this paper we propose a  score of an image to use for coreset selection in image classification and semantic segmentation tasks. The score is the entropy of an image as approximated by the bits-per-pixel of its compressed version.  Thus the score is intrinsic to an image and does not require supervision or training. It is very simple to compute and readily available as all images are stored in a compressed format. The motivation behind our choice of score is that most other scores proposed in literature are expensive to compute. More importantly, we want a score that captures the perceptual complexity of an image.  Entropy is one such measure, images with clutter tend to have a higher entropy.  However sampling only low entropy iconic images, for example,   leads to biased learning and an overall decrease in test performance with current deep learning models. To mitigate the bias we use a graph based method that increases the spatial diversity of the selected samples.  We show that this simple score yields good results, particularly for semantic segmentation tasks.

\end{abstract}    
\section{Introduction}
\label{sec:intro}
Deep learning has made tremendous progress in the past few years exploiting the scale of large training sets, among other factors. Recently data centric methods, such as training on pruned dataset~\cite{sorscher2022beyond}, or using non uniform mixing strategies~\cite{wu2020curricula} have become standard practice for training large scale models. In these methods a score is attached to each data instance, and an instance is selected (or not) for training using the ordered scores of  available instances.  In this paper we focus on data pruning for computer vision tasks where a subset of  the instances available is used for training with minimal loss of performance.

In previous  data pruning approaches for computer vision, mostly  on the classification task, scores naturally reflect the learning task at hand. That is, they are based on the distribution of input and its label(s). This implicitly pays less attention to the input itself. Also modeling this distribution accurately is expensive - usually some steps of training have to be done before the scores can be calculated~\cite{sorscher2022beyond, jiang2020characterizing, paul2021deep, Toneva2018AnES}. Thus one of the key questions, raised in literature, is  how early in training can the instances to be pruned identified~\cite{paul2021deep}?

We start with four observations. First, the vision perception literature recognizes that images with less clutter, simpler background, iconic objects, are processed faster by the human visual system~\cite{rosenholtz2007measuring}. Many measures for characterising human perception of scene complexity have been proposed~\cite{kyle2023characterising}, among which the information theoretic measure of entropy is not only the simplest but  also captures the clutter in an image well~\cite{rosenholtz2007measuring}. 

Second, images with less clutter and a plain background, also lead to better  labeling of pictures by young children~\cite{Gan2008TransferChild}. In general child development literature shows that the nature of picture books for young children  leads to different transfer learning experience. Similar studies for deep learning can yield interesting insights.  

Third, in natural language processing scores such as length of sentences, or word rarity, have been proposed as difficulty scores for machine translation~\cite{platanios2019competence}. Sentences that are shorter are easier to translate than sentences that are longer; here  difficulty is a measure of only the input sentence, not of the input, output sentences.  Machine translation is similar to the semantic segmentation task in computer vision.  

Fourth, it has been shown that complexity of an image (as measured by lossless bits-per-pixel) and the likelihood of generating that image are negatively correlated~\cite{serra2019input}. This provides an alternate method to compute the entropy of an image, that has only been used in context of out of distribution sample detection~\cite{serra2019input}.

Inspired by these observations, we  raise the following questions - 
\begin{itemize}
    \item What is the intrinsic measure of complexity of an image?  
    \item Can a deep learning model learn from the intrinsically more (or less) difficult images in a dataset and  transfer the knowledge effectively to other images in the dataset?
    \item Can this coreset selection and training work for tasks such as image classification and semantic segmentation? Particularly the latter which is less explored task in  literature. 
\end{itemize}

In this paper we propose  the  bits-per-pixel ($BPP_J$) of a JPEG~\cite{Penn92JPEG}  encoded image as a  measure  of the perceptual complexity of the image. Bits-per-pixel of an lossless compressed version of image  is an upper bound for the entropy of an image~\cite{Cover2006}. We acknowledge that JPEG  is not lossless compression, but  in most cases we   do not have access to the raw uncompressed images. We assume that the images available have been stored at high quality in order to avoid compression artifacts, and hence can be approximated as lossless. 
The three dataset that we work on in this paper, CIFAR~\cite{KriCIFAR}, VOC~\cite{EveringhamGWWZ10} and ADE20K~\cite{zhou2019semantic} have in general good quality images, with CIFAR being the dataset with the most variations in quality. We  experiment with alternate methods to estimate entropy in the methods section. 
Also as part of ongoing research we are exploring other perceptual scores~\cite{kyle2023characterising,Wang_2021_CVPR}.

\begin{figure}[t]
  \centering
   \includegraphics[width=\linewidth]{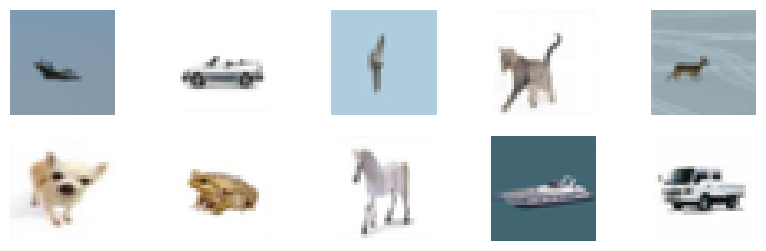}
   \includegraphics[width=\linewidth]{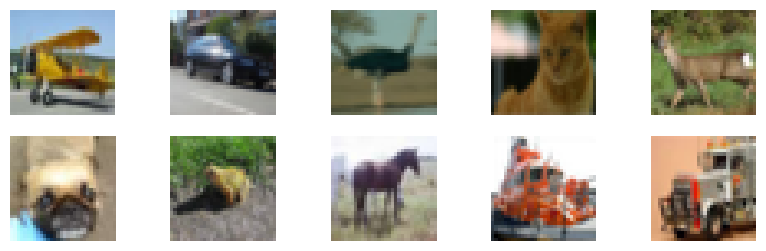}
     \includegraphics[width=\linewidth]{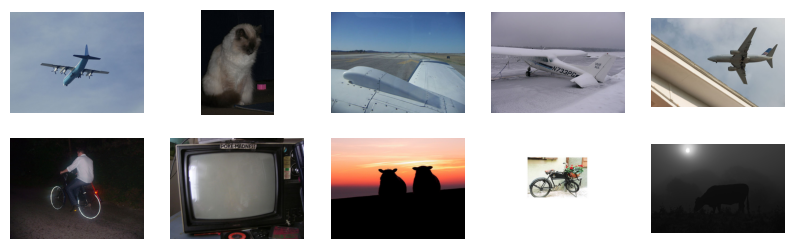}         
       \includegraphics[width=\linewidth]{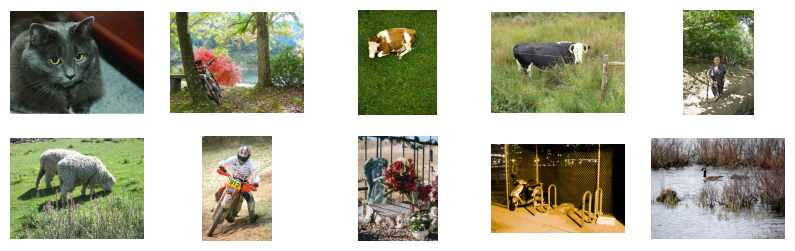}

   \caption{Example of simple images from CIFAR (first two rows), complex  images from CIFAR (next two rows). Simple images from VOC (next two rows), and complex images from VOC (last two rows).
   Simple images are those with lowest $BPP_J$, while complex images have highest $BPP_J$. Note we are using simple and complex terms in line with the motivation of perceptual complexity.}
   \label{fig:iconic}
\end{figure}
\begin{figure}[h]
  \centering
   \includegraphics[width=\linewidth]{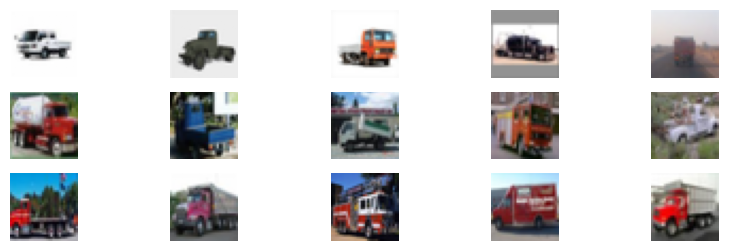}
    \includegraphics[width=\linewidth]{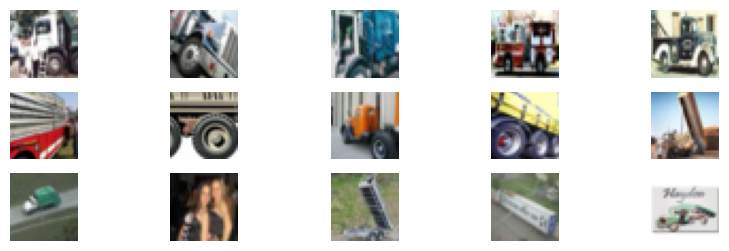}
   \caption{Images from CIFAR. Top row: lowest $BPP_J$, Second row: highest prototypical score~\cite{sorscher2022beyond}. Third row: highest consistency score~\cite{jiang2020characterizing}. In the last three rows the same scores are used but in the reverse order. Perceptually top row shows simple images, fourth row shows complex image. Second, third rows show images that are easy or redundant in training process, while fifth, sixth rows show images that are hard or important for training.}
   \label{fig:simple-complex}
\end{figure}
In Fig.~\ref{fig:iconic}, we show  images from CIFAR~\cite{KriCIFAR} and VOC~\cite{EveringhamGWWZ10} dataset, selected by $BPP_J$ score. Simple CIFAR images  would be good candidates for iconic images in  a child's picture book.
Compared to  other scores for data pruning in computer vision~\cite{jiang2020characterizing,Toneva2018AnES,paul2021deep} (and references therein), our intrinsic score, $BPP_J$, does not use  supervised labels, or even unsupervised cluster distances~\cite{sorscher2022beyond}. In Fig.~\ref{fig:simple-complex} we show images of trucks from CIFAR selected using three different scores. One of the drawbacks of using $BPP_J$ is immediately visible -- images are too self-similar and sample a small subset of the data~\footnote{ shows compression artifacts in CIFAR that affect $BPP_J$ score. The fourth, fifth image in the top row have low $BPP_J$ because they are  highly compressed. }. On the other hand  the hard images as identified by consistency score are the most diverse, in fact  a wrongly labelled image is correctly scored as highly inconsistent.

Less diversity implies that the pruned training set is biased and although the model easily overfits to the  training set, it does not generalize well to the test set (test set is not pruned). This issue of less data diversity, and the resultant bias, in actively sampled, or pruned, training dataset has been raised  in~\cite{sener2017active,zheng2022coverage}. One method proposed to solve it is to use better data coverage sampling methods. 
We use the graph density approach of~\cite{ebert2012ralf} to diversify the sampled dataset. Consequently we are exploiting $BPP_J$, while exploring the sample space using a K-NN graph.
For the semantic segmentation task we also use a novel feature (in this context), the histogram of ground truth labels of an image, and use the Jensen-Shannon divergence distance to build the K-NN graph.  This captures well the semantic similarity of a pair of images. 

%
Sampling methods have also been proposed in~\cite{ortega2018graph} (and references therein). 
Other methods for increasing data diversity, or reducing bias, could be the unbiased approaches proposed in, for example~\cite{lee2021learning,hong2021unbiased}.  These approaches recognize that datasets for training machine learning models tend to be biased unless the data is collected with  care, but they have not explored the case where bias may be introduced knowingly and systematically, for example when using data pruning methods. As part of ongoing research we are exploring these methods.

Our results for image classification task on CIFAR  show that the scoring using $BPP_J$ does not do well on its own, but 
combined with graph based sampling its comparable to  SOTA. We show that using $BPP_J$  and graph based sampling we are able to achieve substantially better results than random pruning for semantic segmentation tasks. We are not aware of other data pruning methods/scores for this task.

\section{Method}
\label{sec:method}

The score $BPP_J$ of an image  is the bits-per-pixel of its JPEG encoded version. We use the byte size of JPEG divided by its dimensions to calculate $BPP_J$. 
As ADE20K and VOC datasets store images as high quality JPEG, this is straightforward. For CIFAR data we use JPEG standard, through OpenCV~\footnote{https://opencv.org/}, at the highest quality setting, to  compress the numpy array of each image. Note that CIFAR  dataset is a subset of Tiny Images which are scrapped images from the web. These images were most likely already lossy compressed. Thus even for CIFAR we do not have access to raw uncompressed images. If raw uncompressed images are available we can use a lossless compression encoder as in~\cite{serra2019input}.  

Another method for estimating the bits-per-pixel of an image is the log likelihood of that image inferred from a trained generative model~\cite{nalisnick2018deep}. In the past likelihood has been used for out-of-distribution detection, but as far as we know not for data pruning. We  use it as a score, $NLL$, to prune data for the image classification task.  It has also been observed~\cite{serra2019input} that generative likelihood of an image is inversely correlated with complexity of the image that is measured by the entropy of a lossless encoder. Subtracting the entropy from log likelihood of an image compensates for the complexity of the image~\cite{serra2019input}. This score, $CPX = NLL-BPP_J$ gives, surprisingly, the best results for the image classification task.  

In ~\cite{sorscher2022beyond} authors  cluster features, inferred from a self-supervised model, of images, using k-means algorithm. The distance to the nearest cluster centroid is the prototypicality score of an image, $PS$. k-means is a method for vector quantization and the distance to the nearest centroid is the distortion incurred in using the quantizer. By rate-distortion theory~\cite{Cover2006} we can consider the distortion to have a bit-per-pixel interpretation under lossy compression conditions. The difference between $PS$ and $BPP_J$ is that the latter uses JPEG which has no learning component. $PS$, on the other hand has a learning component but it is trained on a different dataset in an unsupervised manner. Also the distortion in $PS$ is on the decoded features, and in $BPP_J$ is on the decoded image; in fact its not clear that the features used in $PS$ can be used to decode (generate) the image. On the other hand $NLL$ and $CPX$ use generative model learnt on the dataset itself, and have the capability to generate (decode) the image itself. They are unsupervised generative models. Note that in this work we do not compare with the any other scores for coreset selection, data pruning, because they rely on supervised learning.




To increase data diversity and remove bias arising from self-similar sampled subset, we use the graph density method proposed in~\cite{ebert2012ralf}. A K-NN graph is built, where  each image  is a node, and each node has K edges that connect the  top-K nearest neighbours. Graph is made symmetric and weighted by using a Gaussian kernel on the distance associated with  an edge. Score of an image is attached to corresponding image.  To sample nodes from the graph, iteratively the highest scored node is selected, and its neighboring nodes' scores are down-weighted. Since the distance between two nodes is a representation of their semantic similarity, score of neighboring nodes that are farther away from the selected node are down-weighted relatively less than those of nodes that are closer. This is done  to maximize the diversity of the sampled data and implemented via reverse message passing, where the neighboring nodes receive a weighted message from the selected node and use it to update their score~\cite{ebert2012ralf}\footnote{To select samples in ascending order of scores, the node with the lowest score is selected and its neighbours' score up-weighted}.  

\begin{figure}[t]
  \centering
    \includegraphics[width=\linewidth]{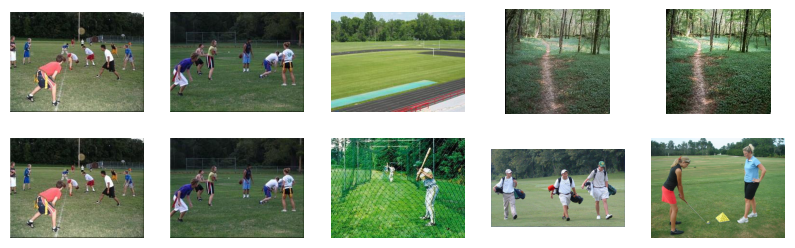}
   \caption{4-nearest neighbours, for image in the first column. Using SWAV features, top row. Using histograms of semantic labels, bottom row. ADE20K dataset.}
   \label{fig:DCvDJS}
\end{figure}

Distance between nodes of the K-NN graph can be defined using features of the images, for example features inferred from self-supervised models such as SWAV~\cite{caron2020unsupervised}. Distance here would be the standard $l_2$ metric. K-NN graphs built this way are denoted as $G_S$. This requires a pre-trained model to infer features from, and though it works reasonably well for image classifcation task, it does not work well for semantic segmentation. 
For the latter we propose  to use the histogram of ground truth labels as the feature, and the Jensen Shannon divergence~\cite{End2003JS} between the histograms as the distance between nodes. K-NN graphs built this way are denoted as $G_H$
In Fig.~\ref{fig:DCvDJS} we see that nearest neighbours using SWAV features have very limited semantic similarity, the third and fourth neighbours on the top row do not have any people in it. 
The biggest advantage of using histograms of labels as features is that  it allows for diversity in the semantic space, not in the feature space. In semantic segmentation where class imbalance is an inherent problem, sampling the semantic space is more suitable; as we will show in results there is a substantial gain in performance using this feature.  It should be noted that having semantic labels implies that annotations are available, while SWAV features do not have that requirement. 

\section{Results}
\label{sec:results}
\small{
\begin{table}[t]
    \centering
    \begin{tabular}{|c|c|c|c|c|}
    \hline
          &  1000&  5000&  12500&  25000  \\ \hline
         $RND$& 58.12 &  80.14&  89.32& 92.12   \\\hline
         $BPP_J$& 34.56 &  57.39&  79.23&   88.22 \\\hline
         $PS$ &52.11  &  78.03&  87.47&    91.58\\\hline
         $NLL$ & 41.09 &  66.72&88.9  & 90.52   \\\hline
         $CPX$ & 57.17 &  80.46&90.01  & 92.31   \\\hline
        $BPP_J+G_S$ & 54.5 & 79.43 & 89.1 & 92.34   \\\hline
    \end{tabular}
    \caption{Accuracy (\%) for CIFAR dataset for different sampling methods (rows) and different pruned sample size (columns). $RND$ uses no random sampling, $BPP_J$, $PS$, $NLL$, $CPX$ use the score only, while $BPP_J+G_S$ uses score and graph based sampling. Best results are obtained by using ascending order for all scores.   }
    \label{tab:res_cifar}
\end{table}}
In this paper we use CIFAR-10 dataset for image classification experiments. We use Resent18 model with hyper-parameters as in~\cite{sorscher2022beyond}. For semantic segmentation we  use VOC and ADE20K dataset, MobileNet model as the encoder, and one convolution along with deep supervision as the decoder ~\cite{zhou2019semantic}.  For ADE20K we use the default hyper-parameters, for VOC we lower the learning rate.  We estimate the likelihood, $NLL$, of an image in CIFAR, using GLOW models~\cite{kingma2018glow,nalisnick2018deep} trained on CIFAR. We use $NLL$ and $BPP_J$ to calculate $CPX$. 

Table~\ref{tab:res_cifar} shows the result for image classification task. $BPP_J$ does not do well at all, and neither does $NLL$. We tried sampling both in descending and ascending order of $BPP_J$ and $NLL$ and the best results we got were with descending order -- that is with high bits-per-pixel  images.  Accounting for complexity of image, $CPX$,  estimated by generative model does well, beating $PS$ and random sampling for lower pruning rates. Also using graph sampling substantially improves performance for $BPP_J$ score, showing the importance of data diversity. 

\begin{figure}[b]
  \centering
    \includegraphics[width=\linewidth]{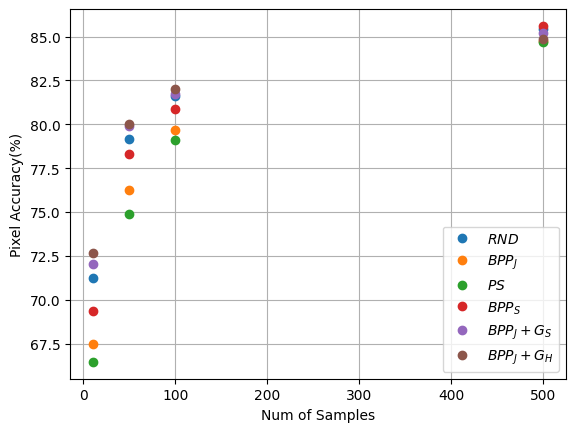}
   \caption{Results for semantic segmentation using VOC data. Legend similar to Table~\ref{tab:res_cifar}, $BPP_J+G_S$ uses $BPP_J$ as score, and $G_S$ as the k-NN graph. $BPP_J+G_H$ uses $G_H$ as the k-NN graph.  }
   \label{fig:resVOC}
\end{figure}

\begin{figure}[b]
  \centering
    \includegraphics[width=\linewidth]{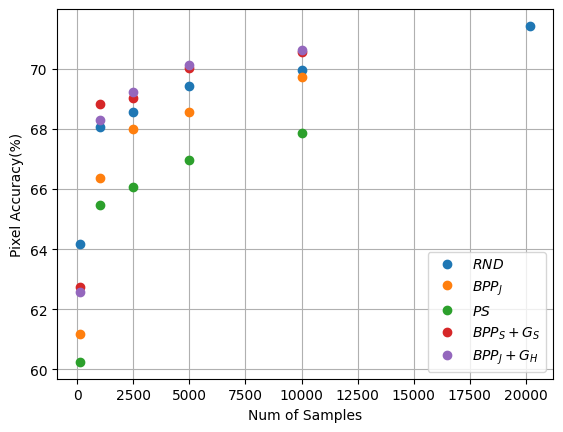}
   \caption{Results for semantic segmenation on ADE20K dataset. Legends same as in Fig.\ref{fig:resVOC}.}
   \label{fig:resADE}
\end{figure}

In Fig.~\ref{fig:resVOC} and Fig.~\ref{fig:resADE} semantic segmentation results for VOC and ADE dataset are shown. For both datasets we see that $BPP_J$ (again in descending order) by itself (without graph sampling) does better than prototypicality score (in ascending order). $BPP_J$ along with k-NN graph defined by histograms of labels $G_H$ give measurable improvement over random sampling and over graph sampling using SWAV featuers $G_S$. $PS$ does poorly for semantic segmentation because it exaggerates the class imbalance problem.

In this paper we  proposed a simple, intrinsic, perceptual complexity score for coreset selection. This score  requires no training and does not use the labels; when we use labels, in $G_H$, we are  not training on them  and coreset selection can still be done before training starts.  One of our original hypothesis of learning from iconic images does not seem, as of now, correct for deep learning models. We also believe that perceptual scores could be used a priors to be updated with label and task information, once training starts. The positive results of this study are  usage of $CPX$ and on semantic segmenation.

{
    \small
    \bibliographystyle{ieeenat_fullname}
    \bibliography{main}

\begin{thebibliography}{26}
\providecommand{\natexlab}[1]{#1}
\providecommand{\url}[1]{\texttt{#1}}
\expandafter\ifx\csname urlstyle\endcsname\relax
  \providecommand{\doi}[1]{doi: #1}\else
  \providecommand{\doi}{doi: \begingroup \urlstyle{rm}\Url}\fi

\bibitem[Caron et~al.(2020)Caron, Misra, Mairal, Goyal, Bojanowski, and Joulin]{caron2020unsupervised}
Mathilde Caron, Ishan Misra, Julien Mairal, Priya Goyal, Piotr Bojanowski, and Armand Joulin.
\newblock Unsupervised learning of visual features by contrasting cluster assignments.
\newblock \emph{Advances in neural information processing systems}, 33:\penalty0 9912--9924, 2020.

\bibitem[Cover and Thomas(2006)]{Cover2006}
Thomas~M. Cover and Joy~A. Thomas.
\newblock \emph{Elements of Information Theory 2nd Edition (Wiley Series in Telecommunications and Signal Processing)}.
\newblock Wiley-Interscience, 2006.

\bibitem[Ebert et~al.(2012)Ebert, Fritz, and Schiele]{ebert2012ralf}
Sandra Ebert, Mario Fritz, and Bernt Schiele.
\newblock Ralf: A reinforced active learning formulation for object class recognition.
\newblock In \emph{2012 IEEE Conference on Computer Vision and Pattern Recognition}, pages 3626--3633. IEEE, 2012.

\bibitem[Endres and Schindelin(2003)]{End2003JS}
D.M. Endres and J.E. Schindelin.
\newblock A new metric for probability distributions.
\newblock \emph{IEEE Transactions on Information Theory}, 49\penalty0 (7):\penalty0 1858--1860, 2003.

\bibitem[Everingham et~al.(2010)Everingham, Gool, Williams, Winn, and Zisserman]{EveringhamGWWZ10}
Mark Everingham, Luc~Van Gool, Christopher K.~I. Williams, John~M. Winn, and Andrew Zisserman.
\newblock The pascal visual object classes (voc) challenge.
\newblock \emph{Int. J. Comput. Vis.}, 88\penalty0 (2):\penalty0 303--338, 2010.

\bibitem[Hong and Yang(2021)]{hong2021unbiased}
Youngkyu Hong and Eunho Yang.
\newblock Unbiased classification through bias-contrastive and bias-balanced learning.
\newblock \emph{Advances in Neural Information Processing Systems}, 34:\penalty0 26449--26461, 2021.

\bibitem[Jiang et~al.(2020)Jiang, Zhang, Talwar, and Mozer]{jiang2020characterizing}
Ziheng Jiang, Chiyuan Zhang, Kunal Talwar, and Michael~C Mozer.
\newblock Characterizing structural regularities of labeled data in overparameterized models.
\newblock \emph{arXiv preprint arXiv:2002.03206}, 2020.

\bibitem[Kingma and Dhariwal(2018)]{kingma2018glow}
Diederik~P. Kingma and Prafulla Dhariwal.
\newblock Glow: Generative flow with invertible 1x1 convolutions, 2018.

\bibitem[Krizhevsky et~al.()Krizhevsky, Nair, and Hinton]{KriCIFAR}
Alex Krizhevsky, Vinod Nair, and Geoffrey Hinton.
\newblock Cifar-10 (canadian institute for advanced research).

\bibitem[Kyle-Davidson et~al.(2023)Kyle-Davidson, Zhou, Walther, Bors, and Evans]{kyle2023characterising}
Cameron Kyle-Davidson, Elizabeth~Yue Zhou, Dirk~B Walther, Adrian~G Bors, and Karla~K Evans.
\newblock Characterising and dissecting human perception of scene complexity.
\newblock \emph{Cognition}, 231:\penalty0 105319, 2023.

\bibitem[Lee et~al.(2021)Lee, Kim, Lee, Lee, and Choo]{lee2021learning}
Jungsoo Lee, Eungyeup Kim, Juyoung Lee, Jihyeon Lee, and Jaegul Choo.
\newblock Learning debiased representation via disentangled feature augmentation.
\newblock \emph{Advances in Neural Information Processing Systems}, 34:\penalty0 25123--25133, 2021.

\bibitem[Nalisnick et~al.(2018)Nalisnick, Matsukawa, Teh, Gorur, and Lakshminarayanan]{nalisnick2018deep}
Eric Nalisnick, Akihiro Matsukawa, Yee~Whye Teh, Dilan Gorur, and Balaji Lakshminarayanan.
\newblock Do deep generative models know what they don't know?
\newblock \emph{arXiv preprint arXiv:1810.09136}, 2018.

\bibitem[Ortega et~al.(2018)Ortega, Frossard, Kova{\v{c}}evi{\'c}, Moura, and Vandergheynst]{ortega2018graph}
Antonio Ortega, Pascal Frossard, Jelena Kova{\v{c}}evi{\'c}, Jos{\'e}~MF Moura, and Pierre Vandergheynst.
\newblock Graph signal processing: Overview, challenges, and applications.
\newblock \emph{Proceedings of the IEEE}, 106\penalty0 (5):\penalty0 808--828, 2018.

\bibitem[Patricia A.~Ganea and DeLoache(2008)]{Gan2008TransferChild}
Megan Bloom~Pickard Patricia A.~Ganea and Judy~S. DeLoache.
\newblock Transfer between picture books and the real world by very young children.
\newblock \emph{Journal of Cognition and Development}, 9\penalty0 (1):\penalty0 46--66, 2008.

\bibitem[Paul et~al.(2021)Paul, Ganguli, and Dziugaite]{paul2021deep}
Mansheej Paul, Surya Ganguli, and Gintare~Karolina Dziugaite.
\newblock Deep learning on a data diet: Finding important examples early in training.
\newblock \emph{Advances in Neural Information Processing Systems}, 34:\penalty0 20596--20607, 2021.

\bibitem[Pennebaker and Mitchell(1992)]{Penn92JPEG}
William~B. Pennebaker and Joan~L. Mitchell.
\newblock \emph{JPEG Still Image Data Compression Standard}.
\newblock Van Nostrand Reinhold, New York, 1992.

\bibitem[Platanios et~al.(2019)Platanios, Stretcu, Neubig, Poczos, and Mitchell]{platanios2019competence}
Emmanouil~Antonios Platanios, Otilia Stretcu, Graham Neubig, Barnabas Poczos, and Tom~M Mitchell.
\newblock Competence-based curriculum learning for neural machine translation.
\newblock \emph{arXiv preprint arXiv:1903.09848}, 2019.

\bibitem[Rosenholtz et~al.(2007)Rosenholtz, Li, and Nakano]{rosenholtz2007measuring}
Ruth Rosenholtz, Yuanzhen Li, and Lisa Nakano.
\newblock Measuring visual clutter.
\newblock \emph{Journal of vision}, 7\penalty0 (2):\penalty0 17--17, 2007.

\bibitem[Sener and Savarese(2017)]{sener2017active}
Ozan Sener and Silvio Savarese.
\newblock Active learning for convolutional neural networks: A core-set approach.
\newblock \emph{arXiv preprint arXiv:1708.00489}, 2017.

\bibitem[Serr{\`a} et~al.(2020)Serr{\`a}, {\'A}lvarez, G{\'o}mez, Slizovskaia, N{\'u}{\~n}ez, and Luque]{serra2019input}
Joan Serr{\`a}, David {\'A}lvarez, Vicen{\c{c}} G{\'o}mez, Olga Slizovskaia, Jos{\'e}~F N{\'u}{\~n}ez, and Jordi Luque.
\newblock Input complexity and out-of-distribution detection with likelihood-based generative models.
\newblock In \emph{The Eighth International Conference on Learning Representations}, 2020.

\bibitem[Sorscher et~al.(2022)Sorscher, Geirhos, Shekhar, Ganguli, and Morcos]{sorscher2022beyond}
Ben Sorscher, Robert Geirhos, Shashank Shekhar, Surya Ganguli, and Ari Morcos.
\newblock Beyond neural scaling laws: beating power law scaling via data pruning.
\newblock \emph{Advances in Neural Information Processing Systems}, 35:\penalty0 19523--19536, 2022.

\bibitem[Toneva et~al.(2018)Toneva, Sordoni, des Combes, Trischler, Bengio, and Gordon]{Toneva2018AnES}
Mariya Toneva, Alessandro Sordoni, R{\'e}mi~Tachet des Combes, Adam Trischler, Yoshua Bengio, and Geoffrey~J. Gordon.
\newblock An empirical study of example forgetting during deep neural network learning.
\newblock \emph{ArXiv}, abs/1812.05159, 2018.

\bibitem[Wang et~al.(2021)Wang, Ke, Talebi, Yim, Birkbeck, Adsumilli, Milanfar, and Yang]{Wang_2021_CVPR}
Yilin Wang, Junjie Ke, Hossein Talebi, Joong~Gon Yim, Neil Birkbeck, Balu Adsumilli, Peyman Milanfar, and Feng Yang.
\newblock Rich features for perceptual quality assessment of ugc videos.
\newblock In \emph{Proceedings of the IEEE/CVF Conference on Computer Vision and Pattern Recognition (CVPR)}, pages 13435--13444, 2021.

\bibitem[Wu et~al.(2020)Wu, Dyer, and Neyshabur]{wu2020curricula}
Xiaoxia Wu, Ethan Dyer, and Behnam Neyshabur.
\newblock When do curricula work?
\newblock \emph{arXiv preprint arXiv:2012.03107}, 2020.

\bibitem[Zheng et~al.(2022)Zheng, Liu, Lai, and Prakash]{zheng2022coverage}
Haizhong Zheng, Rui Liu, Fan Lai, and Atul Prakash.
\newblock Coverage-centric coreset selection for high pruning rates.
\newblock \emph{arXiv preprint arXiv:2210.15809}, 2022.

\bibitem[Zhou et~al.(2019)Zhou, Zhao, Puig, Xiao, Fidler, Barriuso, and Torralba]{zhou2019semantic}
Bolei Zhou, Hang Zhao, Xavier Puig, Tete Xiao, Sanja Fidler, Adela Barriuso, and Antonio Torralba.
\newblock Semantic understanding of scenes through the ade20k dataset.
\newblock \emph{International Journal of Computer Vision}, 127:\penalty0 302--321, 2019.

\end{thebibliography}
}


\end{document}